\documentclass{article}
\usepackage[numbers]{natbib}
\usepackage[dvipsnames]{xcolor}

\usepackage[preprint]{neurips_2019}

\usepackage[utf8]{inputenc} 
\usepackage[T1]{fontenc}    
\usepackage{hyperref}       
\usepackage{url}            
\usepackage{booktabs}       
\usepackage{amsfonts}       
\usepackage{nicefrac}       
\usepackage{microtype}      
\usepackage[utf8]{inputenc}
\usepackage{float}
\usepackage{amsmath,amsfonts,amssymb}
\usepackage{graphics}
\usepackage{graphicx}

\title{Constellation Loss: Improving the efficiency of deep metric learning loss functions for optimal embedding.}

\author{%
  Alfonso Medela\\ 
  Tecnalia\\
  Derio, Spain \\
  \texttt{alfonso.medela@tecnalia.com} \\
  \And
  Artzai Picon \\
  Tecnalia\\
  Derio, Spain \\
  \texttt{artzai.picon@tecnalia.com} \\
}

\begin{document}

\maketitle

\begin{abstract}
Metric learning has become an attractive field for research on the latest years. Loss functions like contrastive loss , triplet loss  or multi-class N-pair loss have made possible generating models capable of tackling complex scenarios with the presence of many classes and scarcity on the number of images per class not only work to build classifiers, but to many other applications where measuring similarity is the key. Deep Neural Networks trained via metric learning also offer the possibility to solve few-shot learning problems. Currently used state of the art loss functions such as triplet and contrastive loss functions, still suffer from slow convergence due to the selection of effective training  samples that has been partially solved by the multi-class N-pair loss by simultaneously adding additional samples from the different classes. In this work, we extend triplet and multiclass-N-pair loss function by proposing the \textit{constellation loss} metric where the distances among all class combinations are simultaneously learned. We have compared our \textit{constellation loss} for visual class embedding showing that our loss function over-performs the other methods by obtaining more compact clusters while achieving  better classification results.
\end{abstract}


\section{Introduction}
\label{sec::introduction}

Distance metric learning approaches \cite{metric_shallow_xing2003distance,metric_shallow_xing_weinberger2006distance_triplet_loss,metric_shallow_xing_davis2007information} work by learning embedding representations that keep close together for similar data points while maintaining them far for dissimilar data points. Among distance metric learning applications we can find face recognition, \cite{liu2017sphereface}, signature verification \cite{bromley1994signature}, authorship verification \cite{du2017siamese}, few-shot learning \cite{koch2015siamese,medela2019} and visual similarity for product design \cite{bell2015learning} among others.

With the popularization of convolutional neural networks \cite{krizhevsky2012imagenet,simonyan2014very}, deep metric learning has been deeply analysed on the last years. Deep metric learning \cite{koch2015siamese,metric_deep_hoffer2015deep,metric_deep_yi2014deep,metric_deep_hu2014discriminative,du2017siamese,sohn2016improved_multiclass_loss} have proven to be effective at learning nonlinear embeddings of the data outperforming existing classical methods. Normally, specific network architectures are trained to minimize an euclidean based loss function where a nonlinear embedding representation is learned to bond together embeddings from similar classes while taking apart embeddings of different classes. 

The definition of appropriate loss functions is crucial for fast convergence and optimal global minimum search \cite{koch2015siamese} and they have received lot of attention on the last years. In this sense, losses such as contrastive loss function \cite{contrastive_loss_chopra2005learning} focuses same-class or different-class pairs are normally used. Triplet loss function \cite{metric_shallow_xing_weinberger2006distance_triplet_loss,DBLP:journals/corr/SchroffKP15} extended contrastive loss by considering a query sample and two additional samples (one positive and one negative). This triplet loss simultaneously enlarges the distances between the embeddings of the query and negative sample while reducing the distance between the positive and query samples. However, these methods suffer from slow convergence and poor local optima \cite{sohn2016improved_multiclass_loss} as, at each update, embeddings are only optimized against one negative class. This was partially solved by the incorporation of the multi-class N-pair loss \cite{sohn2016improved_multiclass_loss}  that generalizes triplet loss by simultaneously optimizing against N-1 negative samples from different classes instead of a single negative class yielding to better performance and faster convergence. 

\begin{figure}[H]
\begin{minipage}[b]{1.0\linewidth}
  \centering
  \centerline{\includegraphics[width=12cm]{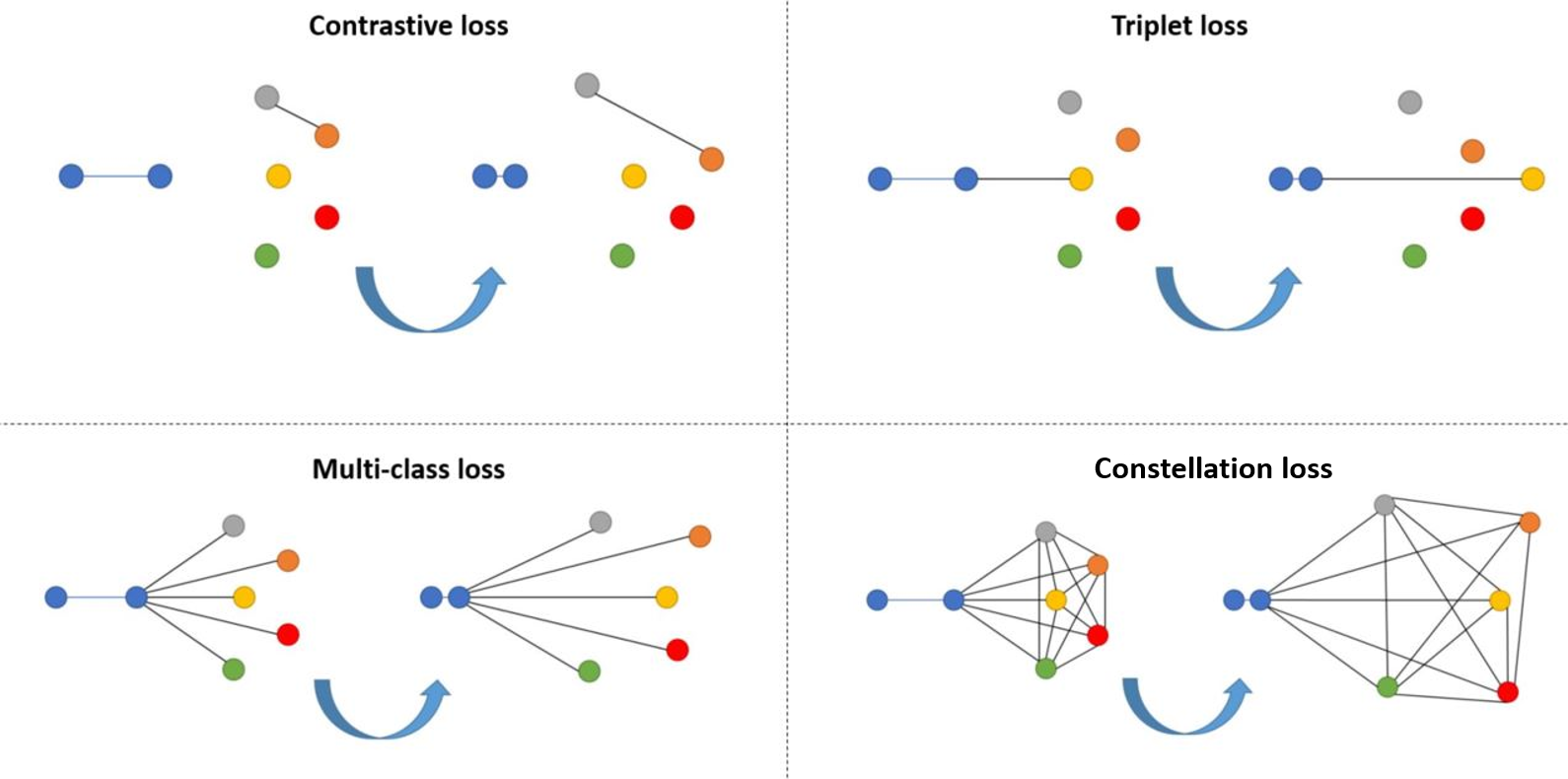}}
\end{minipage}
\caption{Visual representation of a gradient descent step for each of the compared losses: Contrastive loss \cite{contrastive_loss_chopra2005learning}, triplet loss \cite{metric_shallow_xing_weinberger2006distance_triplet_loss},multi-class N-pair loss \cite{sohn2016improved_multiclass_loss} and the proposed Constellation loss}.Each color represent a different class
\label{fig:image_losses}
\end{figure}

However, multi-class N-pair loss function is still ignoring the distances among the different negative classes among them and thus, not assuring optimization among the different negative class embeddings. In this work we extend multiclass-N-pair loss with the proposed \textit{constellation loss} metric where the distances among all class combinations are simultaneously learned. Figure \ref{fig:image_losses} graphically shows the interaction among the different class samples distances during a single gradient descent step for the analyzed losses.

In experiment, we validate that \textit{constellation loss} outperforms other metrics for class embedding tasks resulting into higher class classification performance and better cluster separability metrics such as Silhouete \cite{Rousseeuw1987SilhouettesAG}  and Davis-Boulding index\cite{davies1979cluster} . We also propose an methodology that removes the need of using specific supporting architectures such as Siamese Neural Networks \cite{koch2015siamese} for learning the embedding representations that can be learn by the use of smart batch selection dealing to the same functional cost while improving multi class scalability and reducing training memory needs.

\section{Discriminative loss functions for distance metric learning}
\label{sec::losses}

Traditionally, most of the image classification networks such as AlexNet\cite{krizhevsky2012imagenet}, VGGNet\cite{DBLP:journals/corr/SimonyanZ14a}, GoogleNet\cite{DBLP:journals/corr/SzegedyLJSRAEVR14} or ResNet\cite{DBLP:journals/corr/SchroffKP15} adopted cross-entropy based softmax loss function to solve classical classification problems. However, discriminative metric learning loss functions have better generalization ability \cite{cheng2018deep} and have received more attention for feature learning purposes in the latest years not only for verification problems \cite{liu2017sphereface,metric_shallow_xing_weinberger2006distance_triplet_loss} but also for few-shot learning \cite{koch2015siamese,medela2019} as they overcome learning capabilities of traditional classification approaches under small number of training images conditions. 

This is achieved by learning an image embedding  ${f_i}$ from an image ${x_i}$. This embedding represents a class-representative vector, that is, a vector that contains the most important features associated to the corresponding class of the image. To this end, euclidean-distance-based loss functions are normally used. These functions conceptually constrains the learned embeddings to have distance 0 among elements of the same class and greater distances among elements from different classes.   

To this end, euclidean-distance-based loss functions like contrastive loss \cite{contrastive_loss_chopra2005learning} measure pairs of samples. This was extended by triplet loss \cite{metric_shallow_xing_weinberger2006distance_triplet_loss} by comparing triplets with a positive and a negative samples. Multiclass-N-pair loss objective function \cite{sohn2016improved_multiclass_loss} has focused on improving previous distance metric loss functions by generalizing triplet loss. First, it allows joint comparison  among  more  than  one  negative  examples, concretely, N-1 negative examples and secondly, an efficient batch construction strategy was introduced to reduce computation. This loss function has demonstrated superiority over triplet loss as well as other metric learning functions.  Finally, we propose the constellation loss where distances among all negative classes among them are taken into account simultaneously. The different losses are detailed below.

\subsection{Contrastive loss}

Contrastive loss (\ref{eq:1}) only focuses on positive or negative pairs. Positive pairs are composed by same-class images and negative ones by distinct-class pairs. Formally, the network transforms the pair of input images $\{x_{1,i},x_{2,i}\}$ into $\{f_{1,i},f_{2,i}\}$ embedding vectors. The labels are either $y_i=0$ for positive pairs or $y_i=1$ for negative pairs.

\begingroup
\begin{equation}\label{eq:1}
\begin{split}
\mathcal{L}_{c}=\frac{1}{2N}\sum_{i=1}^{N}[(1-y_i)||f_{1,i} - f_{2,i}||_2^2 + (y_i)\{max(0,m-||f_{1,i} - f_{2,i}||_2)\}^2]
\end{split}
\end{equation}
\endgroup

where $m$ is the margin, usually set to $1.0$ and N is the batch size.
\vspace{0.5cm}

Intuitively, this loss penalizes when a positive pair is far away or a negative pair too close. Therefore, in an optimal case, positives are nearby $0.0$ and negatives close to $1.0$.

\subsection{Triplet loss}

Triplet loss (\ref{eq:2}) goes one step further by taking into account positive and negative pairs at the same time. This is done by setting an anchor, from which a distance will be calculated to a sample of the same class (positive) and a sample of a different class (negative). So, the set of input images is a triplet  $\{x_i^a,x_i^p,x_i^n\}$ and their correspondent embedding vectors are $\{f_i^a,f_i^p,f_i^n\}$. No label is needed.

\begingroup
\begin{equation}\label{eq:2}
\begin{split}
\mathcal{L}_{triplet}=\frac{1}{N}\sum_{i=1}^{N}  max(0,||f_i^a - f_i^p||_2^2 - ||f_i^a - f_i^n||_2^2 +\alpha) 
\end{split}
\end{equation}

where $\alpha$ is a parameter to avoid convergence to trivial solutions and N is the batch size.

The aim of this loss is to maximize the distance between the anchor and the negative whilst minimizing the distance between the anchor and positive. Nonetheless, there is no gain when $||f_i^a - f_i^p||_2^2<||f_i^a - f_i^n||_2^2+\alpha$, and for that reason hard-triplet mining is commonly applied. This technique consist on taking into account only hard or semi-hard triplets, that is, using for computation only the triplets that give a positive loss. This way, it forces the network to try harder and it improves convergence.

\subsection{Multiclass-N-pair loss objective}

Multi-class-N-pair loss objective is a generalization of triplet loss, that incorporates at each optimization update the other negative classes that the triplet loss does not take into account. This allows joint comparison among more than one negative example at each update while reducing the computational burden of evaluating deep embedding vectors. So, when having N classes, the distances to the N-1 negative classes are also considered. When only one negative sample is used for calculation (N=$1$), this loss in comparable to triplet loss.

\begingroup
\begin{equation}\label{eq:4}
\begin{split}
\mathcal{L}_{m-c}=\frac{1}{N}\sum_{i=1}^{N}log(1+\sum_{j\neq i}exp( f_i^{\top}f_j^+-f_i^{\top}f_i^+))
\end{split}
\end{equation}
\endgroup

\subsection{Constellation loss}

The constellation loss takes the best of both triplet and multiclass-N-pair loss. It uses the same batch construction than triplet loss and a similar loss formulation than multiclass-N-pair loss. The hyperparameter K sets the number of triplets we want to incorporate in the formula, this way, taking into account more negative terms than the usual triplet loss. Even though increasing K parameter means a bigger computational effort, we prove for our dataset that at some point the fact of increasing K does not affect much the result. This is due to the randomness in the choice of each term, that can be composed of several distinct negative values. Therefore, there is no need for a high K value to improve triplet loss or multiclass-N-pair loss. The main difference is that multiclass-N-pair-loss substracts dot products of same class pairs whereas constellation loss does something similar to triplet loss by substracting a dot product of an anchor and negative embedding; and a dot product of an anchor and positive embedding.

\begingroup
\begin{equation}\label{eq:5}
\begin{split}
\mathcal{L}_{constellation}=\frac{1}{N}\sum_{i=1}^{N}log(1+\sum_{j}^{K}exp( f_i^{a\top}f_j^n-f_i^{a\top}f_i^p))
\end{split}
\end{equation}
\endgroup

\section{Deep Neural Network for Embedding Learning}
\label{sec::network}

Deep embedding extraction can be performed by the use of a Siamese Neural Network architecture \cite{koch2015siamese,du2017siamese} in a similar way as done in \cite{medela2019}. These network employ a twin network architecture where their weights are tied and joined by a loss function that is computed in the top.  This network is normally selected from the state-of-the-art CNN architectures such as VGGNet \cite{DBLP:journals/corr/SimonyanZ14a}, that was successfully used by \cite{koch2015siamese,medela2019}, Inception \cite{DBLP:journals/corr/SzegedyVISW15}, ResNet\cite{he2016deep}among others.

In our experiments we have chosen Inception V3  \cite{DBLP:journals/corr/SzegedyVISW15} classification architecture, where the last layer is replaced as suggested by \cite{DBLP:journals/corr/SchroffKP15} by a global average pooling layer and a 128-neuron layer as an embedding layer, $f_i$, that acts as a low-dimensional representation of the input images. The embedding layer has a sigmoid activation and it is L2 normalized in the case of triplet and constellation loss. L2 normalization makes the training process more stable for these loss functions. There is no need for L2 normalization when training with multiclass-N-pair objective loss as it already incorporates in the loss. After loss optimization procedure, the trained base network is able to extract a embedding $f_i$ from an image $x_i$ that is able to keep similar samples together and dissimilar one apart.

 \subsection{Smart batch} 
 \label{sec::smartbatch}
 
 However, this siamese neural network approach causes memory and scalability problems in more advanced loss functions where triplets or K-plets are the inputs of the network, specially for large values of $K$. Instead of building a siamese structure we follow an \textit{smart batch} strategy that leads to an equivalent loss formulation which is more optimal in terms of speed, network memory and scalability.
 
 Online triplet mining loss presented in  \cite{DBLP:journals/corr/SchroffKP15} where the authors take the embeddings array, computes all the dot products and euclidean distances, and then selects the hard and semi-hard triplets to compute the loss function. Hard-triplets are where the negative is closer to the anchor than the positive and semi-hard triplets are triplets where the negative is not closer to the anchor than the positive, but which still have positive loss.  
 
 N-pair-mc uses a different batch construction in which two embedding arrays are taken as input. Each array contains one embedding per class and are ordered same way in both arrays. Figure \ref{fig_batch_construct} depicts online batch construction of triplet and N-pair-mc pairs.
 
 

\begin{figure}[H]
\begin{minipage}[b]{1.0\linewidth}
  \centering
  \centerline{\includegraphics[width=10cm]{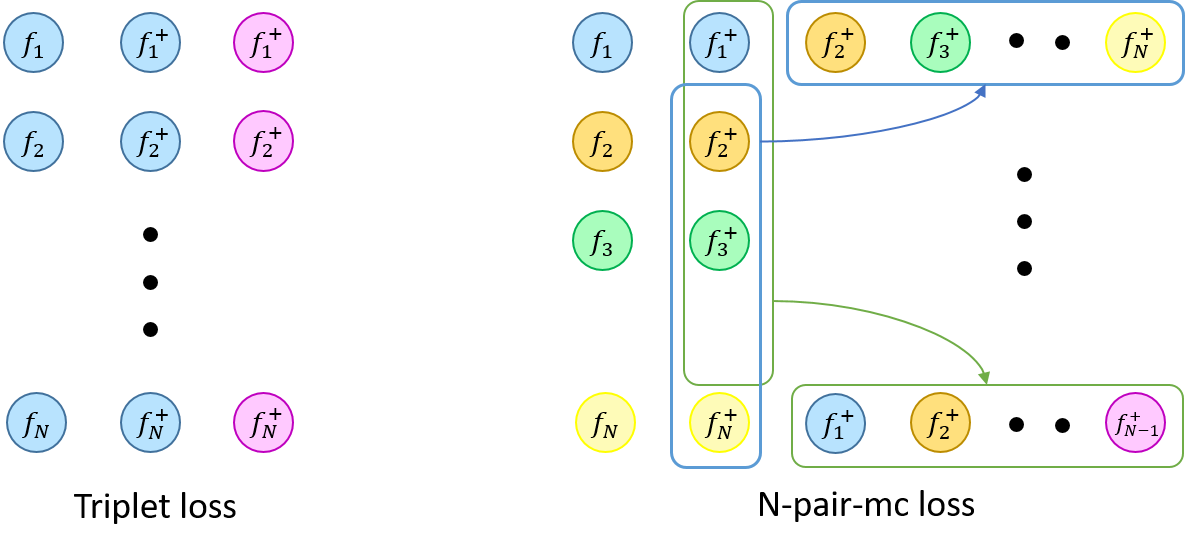}}
\end{minipage}
\caption{Batch construction of triplet loss and N-pair-mc-loss.}
\label{fig_batch_construct}
\end{figure}

In the case of constellation loss, we extend the strategy as in triplet loss but dividing the batch into K groups (see Fig.\ref{fig:constellation_batch}). We later show that even though we use same batch, out loss obtains better results. The nature of constellation loss that takes into account the relationship among all samples in the batch simultaneously constellation loss doesn't need to apply a mask to select hard and semi-hard triplets and neither to compute euclidean distances. It only needs to compute the dot products as triplet loss function does, and apply a mask to find the dot products between same class embeddings, anchor-positive, and the dot products between distinct class embeddings, anchor-negative.

\begin{figure}[H]
\begin{minipage}[b]{1.0\linewidth}
  \centering
  \centerline{\includegraphics[width=10cm]{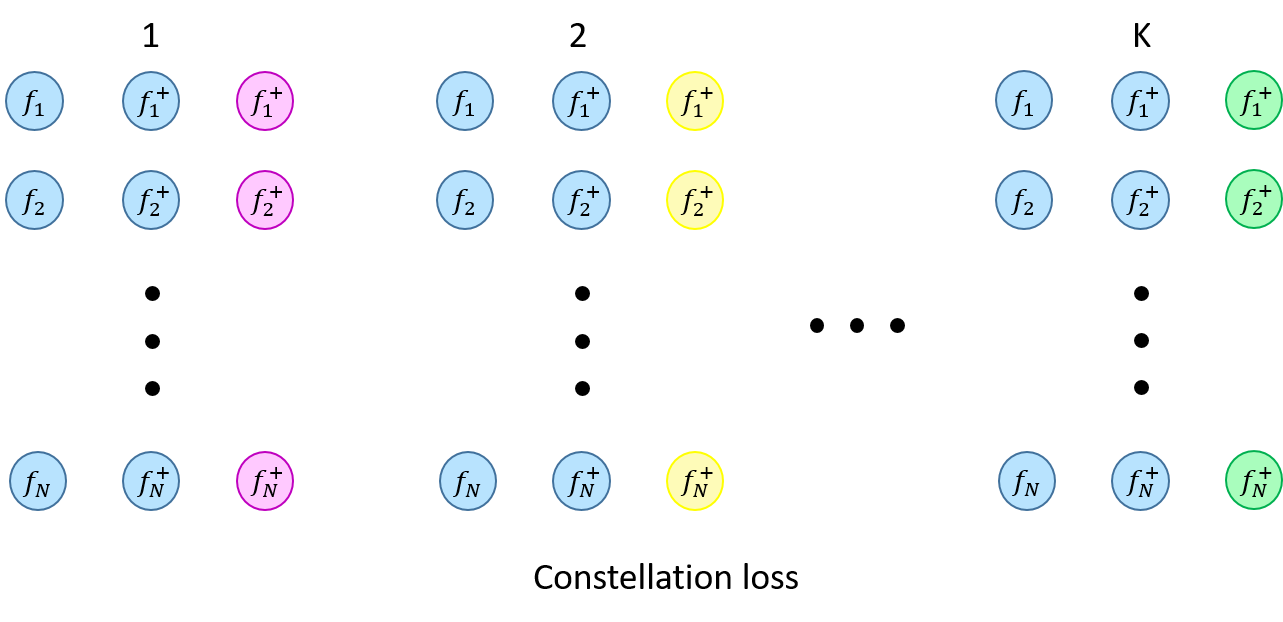}}
\end{minipage}
\caption{Batch construction of constellation loss, based on triplet loss but with K splits.}
\label{fig:constellation_batch}
\end{figure}

\subsection{Validation}
\label{sec::shallowclf}

The proposed backbone network (Inception v3) is trained by means of any of previous loss functions in order to learn the image embeddings $f_i$. This embedding vector is a low-dimensional representation of ${X_i}$ that minimizes its correspondent loss function and thus, are designed to estimate the distance among classes. In order to validate the suitability of these embeddings we analyze, the quality of the class clusters generated by the embedding vector and also the performance of a classification task  that can be achieved by the learned embeddings. This validation procedure is illustrated in figure \ref{fig:SVM} 


\begin{figure}[H]
\begin{minipage}[b]{1.0\linewidth}
  \centering
  \centerline{\includegraphics[width=11cm]{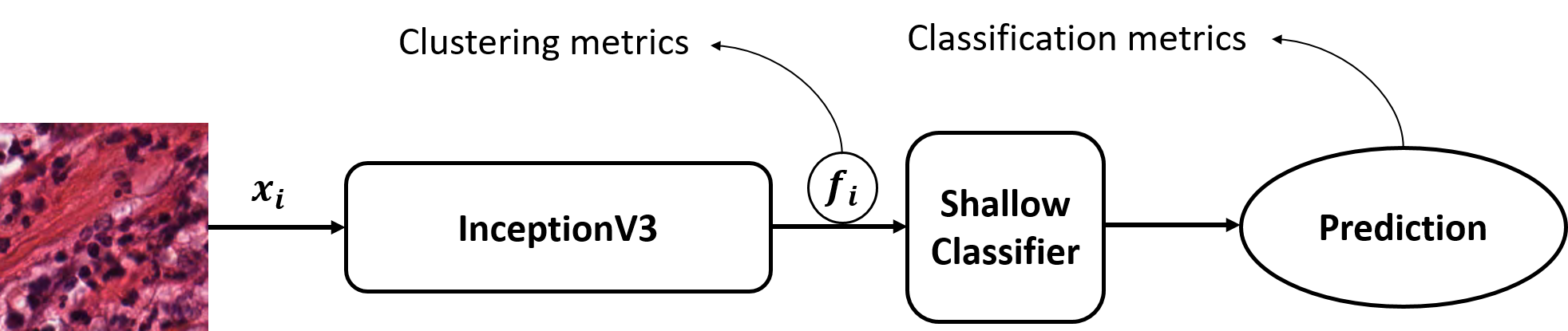}}
\end{minipage}
\caption{Illustration of the validation procedure for the extracted embeddings using a shallow classifier.}
\label{fig:SVM}
\end{figure}

\subsection{Clustering metrics}

We analyse the quality of the class clusters that are generated by the embedding vectors sets to measure the quality of the generated clusters. As the main goal of the network is to create better representations these metrics show how well are the test embeddings grouped in clusters. Two specific metrics were selected:

\textit{Davis-Boulding index}\cite{davies1979cluster} is a metric that evaluates clustering quality. The smaller the value, the better the result. The index uses quantities and features inherent to the dataset and its drawback is that a good value does not imply the best information retrieval.

\textit{Silhouette score}\cite{Rousseeuw1987SilhouettesAG} is a measure of how similar an object is to its own cluster compared to other clusters. The value of this metric ranges from -1 to 1 and the closest to 1 the better the result.

\subsection{Classification metrics}
We also evaluate the classification accuracy that machine learning models can obtain with the learned embeddings. To this end, We select $k$-nearest neighbours as the simpler shallow classifier to predict the class associated to each embedding. First, images go through the network and it outputs the embeddings of all the images, both training and test sets. Then, a $k$-nearest neighbours classifier is used to predict the classes of the test embeddings allowing to measure the obtained accuracy and balanced accuracy metrics.

\section{Experimental Results} 
We assess the performance of the proposed \textit{constellation loss} for visual class embedding extraction. We validate the capability of the tested loss functions to extract appropriate embeddings for the entrusted visual task of histology image classification.

\subsection{Dataset}
Public dataset \cite{kather2016multi} from the University Medical Center Mannheim (Germany). Contains tissue samples obtained from low-grade and high-grade primary tumours of digitalized \textit{colorectal cancer} tissue slides. The database is divided into eight different types of textures that are present on the tumours samples: 1. tumour epithelium, 2. simple stroma, 3. complex stroma, 4. immune cells, 5. debris and mucus, 6. mucosal glands, 7. adipose tissue and 8. background, as depicted in Fig. \ref{fig:nature}. There are 625 image samples per class, producing a total dataset of 5000 image tiles of dimension 150 px x 150 px (74 $\mu$m x 74 $\mu$m).

\begin{figure}[H]
\begin{minipage}[b]{1.0\linewidth}
  \centering
  \centerline{\includegraphics[width=8cm]{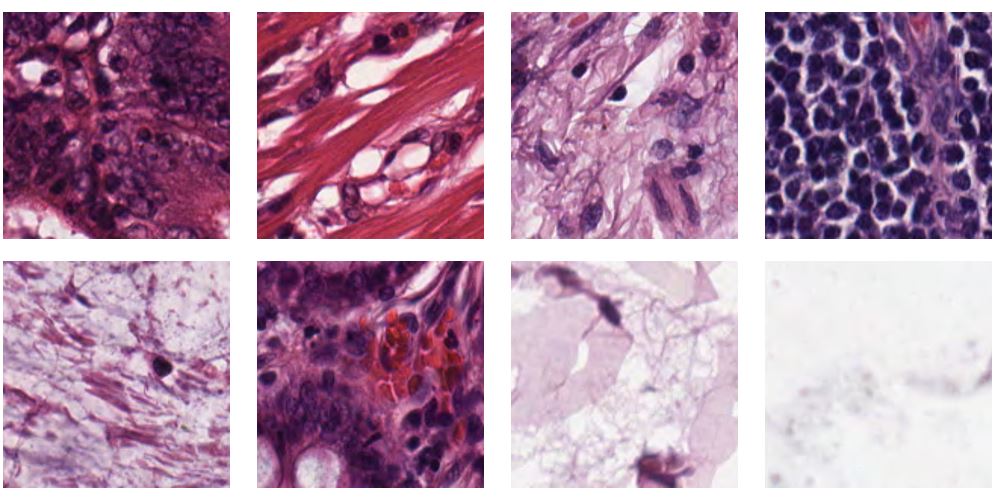}}
\end{minipage}
\caption{Sample images from the dataset. First row: Tumour epithelium, stroma, complex, immune cells. Second row: debris, mucosa, adipose and empty tile samples are depicted.}
\label{fig:nature}
\end{figure}

\subsection{Visual classes embedding extraction}

Our main goal is to evaluate the capabilities of the different loss functions to embed image description. To validate this, we train the deep metric learning architecture detailed in section \ref{sec::network} for the different losses to obtain a network capable of extracting the embedding vector, $f_i$, from an input signal.

\subsubsection{Training}

All the experiments were run for 10 epochs and optimized with Adam with its default parameters. A very simple generator was used in order to feed the images to the network. It makes simple spatial transformations such as inversion in both axes and rotations of multiples of 90 degrees to slightly augment the data. The size of the image is not modified during this process. No more augmentation was considered as the aim of the loss function we are working with is to train on datasets that have not enough images per class.

As a baseline we  trained a classical classification Inception v3 network approach by learning with a softmax loss function that obtained an accuracy of $92.01 \pm 0.99 $\%.

A Inception v3 architecture is trained over the different loss functions following the online smart batch selection as detailed in section ref{sec::smartbatch}.

We ran each network 10 times, each time training and testing in different data splits. We calculated the metrics for each split and gathered them in Table.\ref{tab:table1} by averaging over the splits. This way, we show the mean value and the standard deviation.
We adopted a K-fold cross-validation strategy to validate the constellation loss against the triplet and multiclass-N-pair loss. We divided the data in 10 folds and obtained the metrics for each fold. Then we averaged the results and gathered them in Table.\ref{tab:table1}. The folds are exactly the same for every loss function, with this we mean that they have seen the same train/test split.

\begin{figure}[H]
\begin{minipage}[b]{1.0\linewidth}
  \centering
  \centerline{\includegraphics[width=14cm]{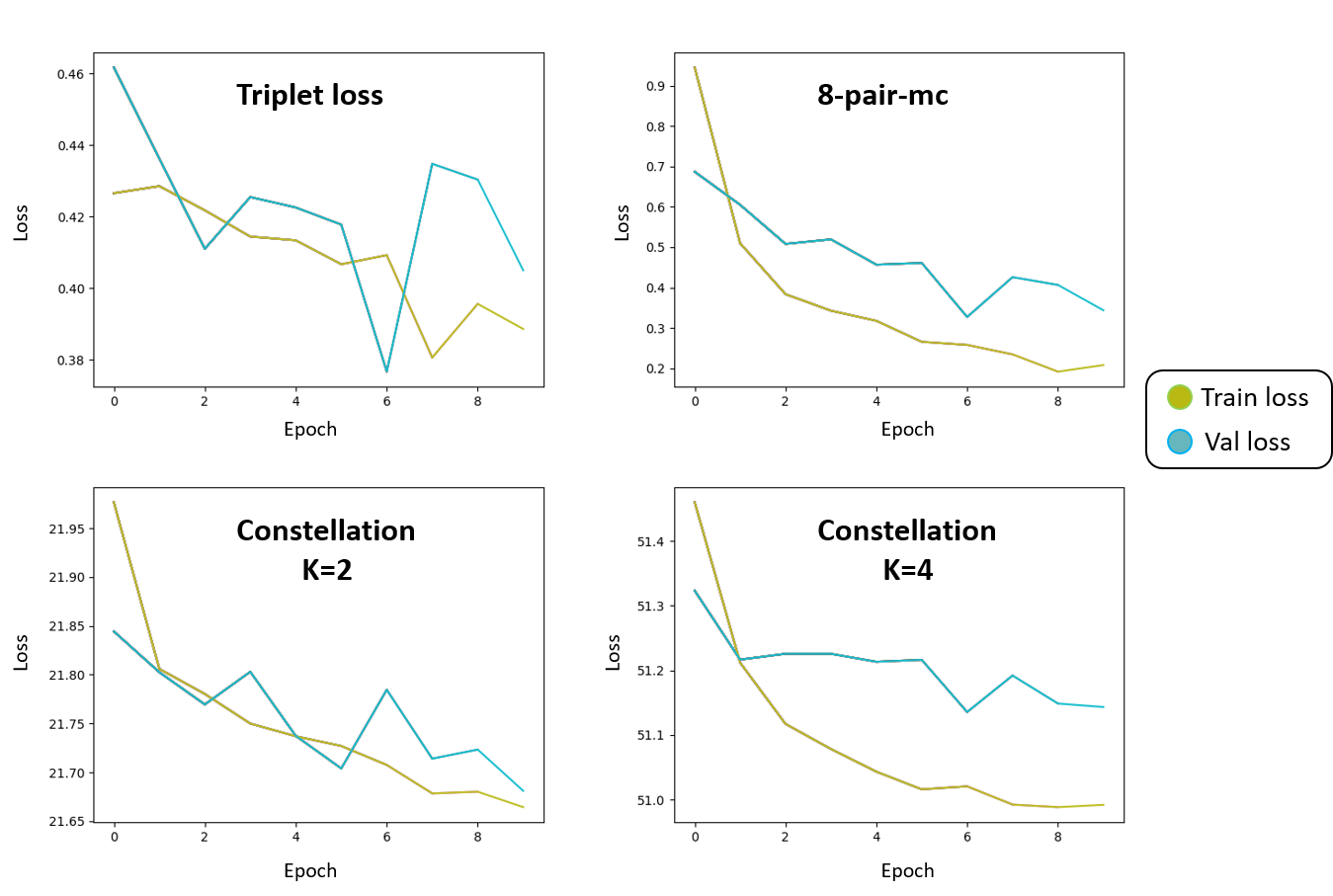}}
\end{minipage}
\caption{Train and validation loss during training. The values of the loss functions are not comparable, the only purpose of the plots is to analyze training loss and validation loss stability and convergence.}
\label{fig:training_loss}
\end{figure}

We observed that triplet loss showed a very unstable evolution during training, whereas the logaritmic loss functions such as multiclass and constellation's convergence was more optimal. They both showed a clear downward direction in train and validation loss alike. Regarding $K$ value, we see that $K=4$ shows signs of overfitting in contrast to $K=2$. However, $K=4$ gives better results in clustering metrics and classification accuracy.

\subsubsection{Results}

The proposed constellation loss beats both triplet and multiclass-N-pair loss in all the metrics. Triplet and 8-pair-mc give Davis-Boulding index values over 0.6, on the other hand, constellation loss gets close to 0.4 for $K=4$. The same happens when we evaluate the loss functions with Silhouettes score, obtaining very similar results for triplet and 8-pair-mc and significantly better for constellation loss with ant $K$. This can also be see in Fig.\ref{fig:tsne}, which shows a 2 dimensional representation of the embedding vectors of the test set. Each color represents a single class. The embedding vector dimension is reduced by the popular t-SNE \cite{maaten2008visualizing} technique, a dimensionality reduction technique that is particularly well suited for the visualization of high-dimensional data. Even if the plots are just for visualization purposes and real data corresponds to a 128-dimensional space, we can notice that the most compact clusters are obtained with constellation loss.

\begin{figure}[H]
\begin{minipage}[b]{1.0\linewidth}
  \centering
  \centerline{\includegraphics[width=14cm]{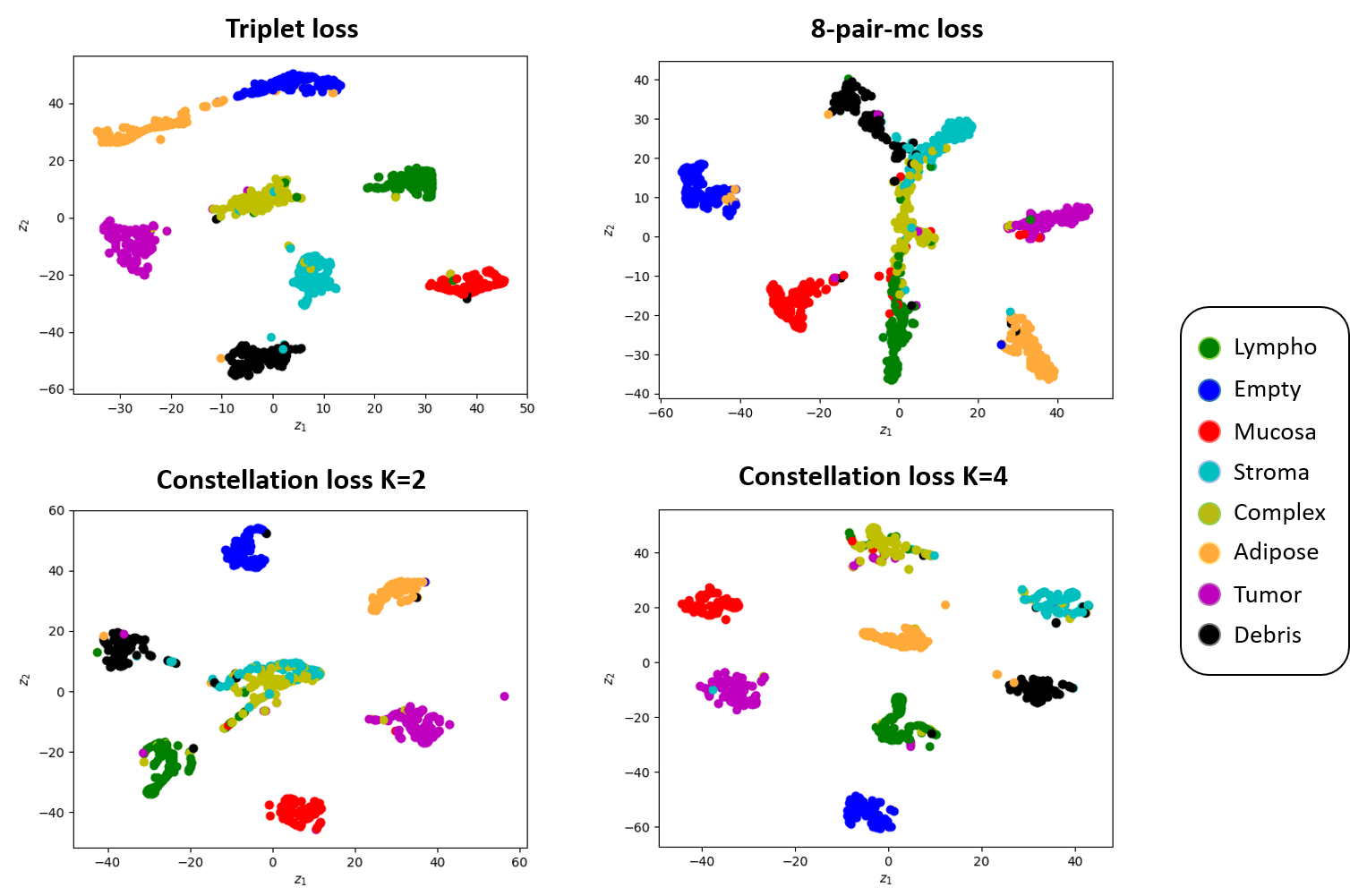}}
\end{minipage}
\caption{Comparison of the clustering capabilities. The figure shows a 2D t-SNE visualization of the embedding vectors in the test set.}
\label{fig:tsne}
\end{figure}

So we can say that the best clusters are obtained with constellation loss for any $K$, based on the clustering metrics. Furthermore, constellation loss shows higher accuracy than triplet and multiclass-N-pair loss for any $K$. In addition, the result we obtain is better than with a classical softmax approach.

\begin{table}[H]
  \begin{center}
    \caption{Mean and standard deviation of the accuracy, BAC, Davis-Bouldin and Silhouette metrics for each experiment. Mean and standard deviation are calculated over the results in each fold.}
    \label{tab:table1}
    \begin{tabular}{l|c|c|c|c}
       & \multicolumn{4}{c}{\textbf{ Metric}}\\
      \hline
      \textbf{Loss} & \textbf{Accuracy} &  \textbf{BAC}  &  \textbf{Davis-Bouldin}  &  \textbf{Silhouette} \\
      \hline
      Triplet & $91.74 \pm 0.68$ & $92.86 \pm 0.72$ & $0.6384 \pm 0.0773$ & $0.6265 \pm 0.0245$\\
      \hline
      8-pair-mc & $91.48 \pm 0.92$  & $91.53 \pm 0.90$  & $0.6037 \pm 0.0386$ & $0.5996 \pm 0.0185$ \\
      \hline
      Constellation k=2 & \textbf{$92.20 \pm 0.65$} & $92.28 \pm 0.55$ &$0.5535 \pm 0.1419$ & $0.7327 \pm 0.0478$\\
      Constellation k=3 & $92.83 \pm 0.73$ & \textbf{$92.86 \pm 0.79$} &\textbf{$0.4558 \pm 0.0518$}& \textbf{$0.7709 \pm 0.0205$}\\
      Constellation k=4 & \textbf{$92.75 \pm 0.39$} & $92.81 \pm 0.44$ &$\textbf{0.4218} \pm \textbf{0.0589}$ & $0.7864 \pm 0.0212$\\
      Constellation k=5 & \textbf{$92.78 \pm 0.68$} & $92.86 \pm 0.69$ &$0.4605 \pm 0.0778$ & $0.7817 \pm 0.0178$\\
      Constellation k=6 & \textbf{$\textbf{92.96} \pm \textbf{0.51}$} & $\textbf{93.01} \pm \textbf{0.56}$ &$0.4373 \pm 0.0511$ & $\textbf{0.7905} \pm \textbf{0.0134}$\\
      Constellation k=7 & \textbf{$92.79 \pm 0.68$} & $92.86 \pm 0.65$ &$0.4334 \pm 0.0546$ & $0.7820 \pm 0.0161$\\
    \end{tabular}
  \end{center}
\end{table}


\subsection{Computing infrastructure}

The experiments were ran on a Gigabyte GeForce GTX Titan X 12GB GDDR5 GPU. The GPU is installed in a local server that we access by an SSH client. We used an Anaconda distribution and the main libraries we used are Keras, Tensorflow and Scikit-Learn. Keras was used for the main architecture and Tensforflow for the loss functions. Scikit-Learn was very helpful for machine learning models like $k$-nearest neighbours, and also for metrics BAC, David-Bouldin index and Silhouette.

\section{Conclusions}

In this work we have compared the performance of different metric learning losses for extracting discriminative embeddings under deep metric learning approach. Our proposed the \textit{constellation loss} metric takes into account the distances among all class combinations are simultaneously learned.
The extracted embeddings have been validated for image classification task over the selected dataset, showing that our loss function over-performs the other methods by obtaining more optimal clusters. This better representation of the classes allows machine learning models such as  $k$-nearest neighbours achieve better results in classification.

Moreover, we showed that constellation loss its more stable than triplet during training and at the same time, surpasses multiclass-N-pair loss in classification accuracy by using a similar mathematical formulation. 
\subsubsection*{Source code}
Source code is available at:
\url{https://git.code.tecnalia.com/comvis_public/piccolo/constellation_loss/}

\subsubsection*{Acknowledgments}

This study has received funding from the European Union's Horizon 2020 research and innovation programme under grant agreement No. 732111 (PICCOLO project \url{https://www.piccolo-project.eu/})


\small
\bibliographystyle{unsrt}
\bibliography{neurips_2019}
\end{document}